\documentclass[sigconf,natbib=true,anonymous=false]{acmart}

\usepackage{comment}
\usepackage{caption}
\usepackage{subcaption}
\usepackage{graphicx}
\usepackage{capt-of}
\usepackage{enumitem}
\usepackage{adjustbox}

\usepackage{amssymb}
\usepackage{pifont}
\usepackage{latexsym}
\usepackage{graphicx}
\usepackage{enumitem}
\usepackage{kantlipsum}
\usepackage{amsmath}
\usepackage{caption}
\usepackage[T1]{fontenc}
\usepackage[utf8]{inputenc}
\usepackage{microtype}
\usepackage{array, makecell}
\usepackage{multirow}
\usepackage{arabtex}
\usepackage{utf8}
\setcode{utf8}
\usepackage{xcolor}
\usepackage{enumitem}
\usepackage{booktabs}
\usepackage{xspace}
\usepackage{multirow}
\usepackage{subcaption}
\usepackage{caption}
\usepackage{stackengine}
\newcolumntype{P}[1]{>{\centering\arraybackslash}p{#1}}
\newcolumntype{C}{>{\centering\arraybackslash\leavevmode}p{\ColWidthNormal}}
\newcommand\datasetName{ArabicaQA}

\def\delequal{\mathrel{\ensurestackMath{\stackon[1pt]{=}{\scriptstyle\Delta}}}}
\definecolor{myCustomColor}{RGB}{58, 121, 164}
\usepackage{pgf-pie}
\usepackage{pgfplots}
\usepackage{tikz}
\usetikzlibrary{shapes}

\pgfplotsset{compat=1.18} 
\usetikzlibrary{patterns}

\usepackage{import}
\usepackage{adjustbox}

\usepackage[disable]{todonotes}
\usepackage{caption}
\usepackage{multirow}
\usepackage{subcaption}
\usepackage{makecell}
\setcellgapes{1pt}
\usepackage{listings}
\usepackage{pifont}

\AtBeginDocument{%
  \providecommand\BibTeX{{%
    \normalfont B\kern-0.5em{\scshape i\kern-0.25em b}\kern-0.8em\TeX}}}

\setcopyright{acmcopyright}
\copyrightyear{2024}
\acmYear{2024}

\acmConference[SIGIR '24]{International Conference on Information Retrieval (SIGIR)}{July 14--18, 2024}{Washington, USA}
\acmBooktitle{SIGIR 2024}

\begin{document}

\title{ArabicaQA: A Comprehensive Dataset for Arabic Question Answering}

\author{Abdelrahman Abdallah}
\email{Abdelrahman.Abdallah@uibk.ac.at}
\affiliation{%
  \institution{University of Innsbruck}
  \streetaddress{Innrain 52}
  \city{Innsbruck}
  \country{Austria}
}
\author{Mahmoud Kasem}
\email{mahmoud.salah@aun.edu.eg}
\affiliation{%
  \institution{ Assuit University}
  \city{Assuit}
  \country{Egypt}}
  
\author{Mahmoud Abdalla}
\email{mahmoud.abdallah@discoapp.ai}
\affiliation{%
  \institution{DISCO AI}
  \streetaddress{Ard El-Marwaha, El-Katamey,Cairo, 11936,}
  \city{Cairo}
  \country{Egypt}
}

\author{Mohamed Mahmoud}
\email{mohamedabokhalil@aun.edu.eg}
\affiliation{%
  \institution{Assuit University}
  \city{Assuit}
  \country{Egypt}
}

\author{Mohamed Elkasaby}
\email{m.abdallah@discoapp.ai}
\affiliation{%
  \institution{DISCO AI}
  \streetaddress{Ard El-Marwaha, El-Katamey,Cairo, 11936,}
  \city{Cairo}
  \country{Egypt}
}

\author{Yasser Elbendary}
\email{yelbendary@discoapp.ai}
\affiliation{%
  \institution{DISCO AI}
  \streetaddress{Ard El-Marwaha, El-Katamey,Cairo, 11936,}
  \city{Cairo}
  \country{Egypt}
}

\author{Adam Jatowt}
\email{adam.jatowt@uibk.ac.at}
\affiliation{%
  \institution{University of Innsbruck}
  \streetaddress{Innrain 52}
  \city{Innsbruck}
  \country{Austria}
}

\renewcommand{\shortauthors}{ Abdallah, et al.}

\begin{abstract}
  In this paper, we address the significant gap in Arabic natural language processing (NLP) resources by introducing ArabicaQA, the first large-scale dataset for machine reading comprehension and open-domain question answering in Arabic. This comprehensive dataset, consisting of 89,095 answerable and 3,701 unanswerable questions created by crowdworkers to look similar to answerable ones, along with 
  additional labels of open-domain questions
  marks a crucial advancement in Arabic NLP resources. We also present AraDPR, the first dense passage retrieval model trained on the Arabic Wikipedia corpus, specifically designed to tackle the unique challenges of Arabic text retrieval. Furthermore, our study includes extensive benchmarking of large language models (LLMs) for Arabic question answering, critically evaluating their performance in the Arabic language context. In conclusion, ArabicaQA, AraDPR, and the benchmarking of LLMs in Arabic question answering offer significant advancements in the field of Arabic NLP. The dataset and code are publicly accessible for further research\footnote{\url{https://github.com/DataScienceUIBK/ArabicaQA}}.

\end{abstract}

\settopmatter{printacmref=true, printccs=true, printfolios=true}


\maketitle

\section{Introduction}

\begin{figure}[ht]
\centering
\begin{subfigure}[b]{0.70\linewidth}
    \centering
    \includegraphics[width=\linewidth]{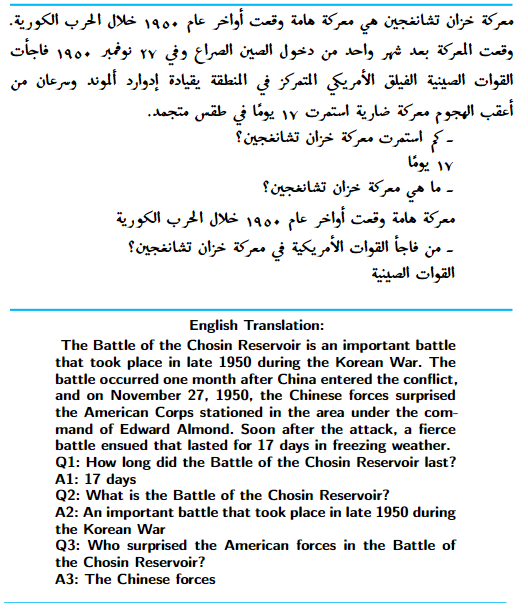}
\end{subfigure}
\caption{An example from the \datasetName{} dataset illustrating a passage about the Battle of Chosin Reservoir during the Korean War, along with corresponding question-answer pairs.}
\label{fig:example}
\end{figure}

Question Answering (QA) has been a central area of study in the field of Natural Language Processing (NLP) for many years. The goal of QA is to design systems that can accurately answer questions posed in natural language. There are several types of QA systems, each with its unique challenges and requirements. For instance, factoid QA systems~\cite{soricut2006automatic,abdallah2023exploring,abdallah2020automated} aim to answer factual questions, typically by retrieving information from a structured knowledge base. 
On the other hand, non-factoid QA systems~\cite{bolotova2022non,bolotova2023wikihowqa} deal with more complex questions that require reasoning or inference. Recent years have seen significant advancements in deep learning, particularly in machine translation~\cite{singh2017machine}, speech recognition~\cite{kamath2019deep}, Question Answering~\cite{allam2012question}, and image recognition~\cite{wu2015image}. These strides are largely due to the development of neural methods and the availability of large datasets for training. Machine Reading Comprehension (MRC)~\cite{nguyen2016ms} and open-domain question answering (Open-domain QA)~\cite{yang2015wikiqa} have emerged as key areas of interest. MRC involves designing systems that can understand a passage and answer questions related to it, while Open-domain QA aims to answer factual questions from a large knowledge corpus without any explicit evidence given.

Arabic, being one of the most widely spoken languages in the world, has unique linguistic features that pose challenges for NLP and IR tasks~\cite{darwish2021panoramic,abdallah2023amurd,bashir2023arabic}. These include rich morphological variations, the use of diacritics, and the existence of Modern Standard Arabic and various dialects. Despite these challenges, Arabic NLP has still not received as much attention as other main languages, leading to a significant gap in resources and research.

In particular, Arabic NLP has seen a noticeable lack of representation in question-answering datasets suitable for MRC and Open-domain QA. This has prevented the development of effective QA systems in Arabic, a language known for its unique linguistic features. In contrast, English has seen significant progress, driven by extensive datasets such as the Stanford QA Dataset (SQuAD)~\cite{rajpurkar2016squad}, Natural Question(NQ)~\cite{kwiatkowski2019natural}, and WikiQA~\cite{yang2015wikiqa}. Our research introduces two major innovations to bridge the gap in Arabic NLP: ArabicaQA and AraDPR. ArabicaQA is a novel dataset for Arabic MRC and Open-domain QA, consisting of 89,095 answerable and 3,701 unanswerable questions for MRC. 76,266 question-answer pairs are also open domain. This makes ArabicaQA the most comprehensive dataset of its kind for Arabic.

Figure \ref{fig:example} illustrates an example of the question-answer pairs derived from an example passage in our dataset, showing the complex storyline structures and rich content that our dataset includes. The creation of ArabicaQA involved a annotation process to ensure high-quality and relevant question-answer pairs. For the MRC, annotators formulate questions based on Wikipedia passages and select the corresponding answers directly from the passages. A second annotator reviews every question and answer, providing a layer of verification and quality control. In the open-domain section, annotators identify questions suitable for Open-domain QA and categorize them based on whether they require a short or long answer, a critical distinction in developing robust Open-domain QA systems.

As our second contribution, we also introduce AraDPR, the first dense passage retrieval model trained on the Arabic Wikipedia corpus. It leverages the power of pre-trained transformers such as BERT and AraBERT~\cite{devlin2018bert,antoun2020arabert}, which is an active area of research in the field of information retrieval~\cite{zhang2023toward}. AraDPR adopts a Bi-encoder architecture~\cite{vaswani2017attention}, allowing for the offline computation of document representations. This transforms the document ranking task into a nearest-neighbor search problem, given the query vector representation. The Bi-encoder architecture offers an attractive alternative to the cross-encoder architecture, where queries and documents are concatenated and fed into a transformer directly. While cross-encoders are practical in a reranking setup, processing candidates generated by the first-stage retrieval system, Bi-encoders directly support single-stage retrieval.

We not only introduce AraDPR but also conduct an extensive benchmarking of Large Language Models (LLMs) for Arabic QA. Our comprehensive benchmarking of LLMs like GPT-3~\cite{brown2020language}, Llama~\cite{touvron2023llama}, and Falcon~\cite{falcon40b} provides several insights into their performance in the Arabic context. These LLMs, known for their remarkable capabilities in understanding and generating text based on context, can exhibit varying performance depending on the language and the specific task at hand. Our findings from this benchmarking exercise offer guidance for researchers and practitioners working on Arabic QA, helping them choose the most suitable models. 

In summary, our contributions are as follows:
\begin{enumerate}
\item \textbf{ArabicaQA Dataset:}  We introduce the first large-scale dataset for Arabic MRC and Open-domain QA, comprising both answered and unanswered questions, along with a distinct open-domain component.
\item \textbf{AraDPR:} We develop the first Arabic dense passage retrieval model trained specifically for Arabic text, addressing the unique challenges of Arabic language retrieval.
\item \textbf{Benchmarking of LLMs:} We provide a comprehensive evaluation of LLMs in Arabic QA, establishing a baseline for their performance in Arabic NLP.
\end{enumerate}

\begin{table*}[t]

  \caption{Comparison of related datasets.}
  \label{tab_compareddataset}
  \centering
  \begin{adjustbox}{width=.8\textwidth,center}
  \begin{tabular}{lllllccc}
  \hline
  \textbf{Dataset} & \textbf{\#Questions} & \textbf{Answer Type} & \textbf{Question Source} & \textbf{Corpus Source}  & \textbf{Language}& \textbf{QA Type} & \textbf{Unanswerable}\\ 
  
  \hline
  
  \makecell[l]{MS MARCO \cite{nguyen2016ms}} & 1M &  \makecell[l]{Generative, \\ Boolean} & Query logs  & Web documents  & English & ODQA & No\\ 
  
  \makecell[l]{SQuAD 1.1 \cite{rajpurkar2016squad}} & 108K & Extractive  & Crowd-sourced & Wikipedia  & English& MRC & No\\ 

  \makecell[l]{SQuAD 2.0 \cite{rajpurkar2018know} }& 158K & Extractive  & Crowd-sourced & Wikipedia  & English & MRC & Yes \\ 

  \makecell[l]{NaturalQuestions \cite{kwiatkowski2019natural}} & 323K &  \makecell[l]{Extractive, \\ Boolean}  & Query logs & Wikipedia  & English & ODQA & Yes \\ 

  \makecell[l]{CNN/Daily Mail \cite{nallapati2016abstractive}} & 1M & Cloze  & \makecell[l]{Automatically \\ Generated} & News  & English & MRC & No \\ 

  \makecell[l]{NewsQA \cite{trischler2016newsqa}} & 119K & Extractive  & Crowd-sourced & News  & English & MRC & Yes \\ 

  ArchivalQA \cite{wang2022archivalqa} & 532K & Extractive & \makecell[l]{Automatically \\ Generated} & News  & English & ODQA & No \\
  \hline
  
  ParSQuAD\makecell[l]{ \cite{9443126}} & 65k &  Extractive &   Translated SQuAD & Wikipedia  & Persian  & MRC & No \\
  PersianQuAD\makecell[l]{ 
  \cite{9729745,mozafari2022peransel}} & 20k &  \makecell[l]{Extractive} & Crowd-sourced & Wikipedia  & Persian  & ODQA & No\\

  \hline

  Arabic-SQuAD \cite{mozannar-etal-2019-neural} & 48,344 & Extractive & \makecell[l]{Translated  \\ SQuAD} & Wikipedia  & Arabic & MRC & No \\
  ARCD\cite{mozannar-etal-2019-neural} & 1,395 & Extractive & \makecell[l]{Crowd-sourced} & Wikipedia  & Arabic & MRC & No \\
  ArabiQA \cite{benajiba2007implementation} & 200 & Extractive & \makecell[l]{Crowd-sourced} & Wikipedia  & Arabic & MRC & No \\
  DefArabicQA \cite{trigui2010defarabicqa} & 50 & Extractive & \makecell[l]{Crowd-sourced} & \makecell[l]{Wikipedia and \\  Googles search }   & Arabic& ODQA & No \\
  \makecell[l]{Translated TREC \\ and CLEF}\cite{abouenour2010evaluated} & 2,264 & Extractive & \makecell[l]{Translated} & \makecell[l]{news articles\\ scientific papers\\ and Web pages.}   & Arabic & ODQA & No \\
  DAWQUAS \cite{ismail2018dawqas} & 3,205 & Extractive & \makecell[l]{Automatically \\ Generated} & \makecell[l]{Web documents}  & Arabic & ODQA & No \\
  QArabPro \cite{akour2011qarabpro} & 335 & Extractive & Crowd-sourced & \makecell[l]{Wikipedia}  & Arabic & MRC & No \\
  \hline
    \textbf{\datasetName}& \textbf{87K}&\textbf{Extractive}  &  \textbf{Crowd-sourced }& \makecell[l]{\textbf{Wikipedia} } & \textbf{Arabic} & \textbf{MRC\&ODQA} & \textbf{Yes} \\
  \hline
\end{tabular}
\end{adjustbox}
\label{tab:dataset}
\end{table*}

\begin{table}[h]
\centering
\caption{Basic Statistics of ArabicaQA.}
\begin{adjustbox}{width=0.4\textwidth,center}
\begin{tabular}{lccc}
\hline
\textbf{Dataset} & \textbf{Training Set} & \textbf{Development Set} & \textbf{Test Set} \\ \hline
MRC (with answers) & 62,186 & 13,483 & 13,426 \\ 
MRC (unanswerable) & 2,596 & 561 & 544 \\ 
Open-Domain\textsuperscript{‡} & 62,057 & 13,475 & 13,414 \\ 
Open-Domain\textsuperscript{†} & 58,676 & 12,715 & 12,592 \\ \hline

\end{tabular}
\end{adjustbox}
\label{datatset:distribution}
\end{table}

\section{Related Work}
\subsection{Open domain QA Benchmarks}
The field of open-domain QA has seen significant advancements in recent years, largely due to the introduction of various QA benchmarks. 
One of the pioneering datasets in this field was the Stanford QA Dataset (SQuAD) 1.1~\cite{rajpurkar2016squad}. SQuAD 1.1 offers 108K question-answer pairs derived from Wikipedia articles. The dataset has been instrumental in the development of advanced Machine Reading Comprehension (MRC) models. Its extension, Open-SQuAD, has further contributed to the field by incorporating unanswerable questions, thereby enhancing the complexity of the tasks and pushing the boundaries of what QA systems can achieve. NarrativeQA~\cite{kovcisky2018narrativeqa}, has taken a distinct approach by using summaries of movies and books to create its question-answer pairs. This has diversified the types of datasets available for training and evaluating QA systems, allowing exploration of different types of questions and answers.

The use of search engine query logs has also been a key development in the field. Datasets like MS MARCO~\cite{nguyen2016ms}, HotpotQA~\cite{yang2018hotpotqa}, and Natural Questions (NQ)~\cite{kwiatkowski2019natural} use queries from Bing and Google and accompany web documents and Wikipedia pages as evidence. 
The emergence of datasets like CNN/Daily Mail~\cite{nallapati2016abstractive}, WhoDidWhat~\cite{onishi2016did}, and ReCoRD~\cite{zhang2018record} has propagated cloze-style tasks, which differ from traditional QA formats. These tasks involve filling in the blanks in a given text, expanding the scope of tasks that QA systems can handle. NewsQA~\cite{trischler2016newsqa}, another notable dataset, focuses on MRC sourcing answers from CNN news articles. This marked a category of temporal document collections in MRC, which allow QA systems to handle time-sensitive questions related to past news articles. ArchivalQA~\cite{wang2022archivalqa} dataset stands out here in terms of scale and scope. It covers a wider and older range of news articles and is designed specifically 
to handle questions related to historical events and long-term trends, providing a unique challenge for QA systems.

In the context of Arabic datasets, Shaheen et al.~\cite{shaheen2014arabic} and \citet{atef2020aqad} have made significant advancements. Shaheen et al.’s work includes an analysis of system components like question analysis and passage retrieval, emphasizing the unique linguistic features of Arabic and the need for specialized approaches in this field. AQAD developed by~\citet{atef2020aqad} is designed to cover Arabic question-answer pairs for MRC. It consists of 17,911 question-answer pairs extracted from 3,381 paragraphs from 299 Wikipedia articles, with 35\% of the questions being unanswerable.
\subsection{Machine Reading Comprehension datasets}
The use of data to enhance reading comprehension has a long history, beginning with the work of~\citet{hirschman1999deep}. They developed a dataset consisting of 600 questions from reading comprehension tests typically administered to students between the 3rd and 6th grades. This initial effort, which used a pattern-matching baseline, was later improved through a rule-based system~\cite{riloff2000rule}. MCTest dataset~\cite{richardson2013mctest}, which was created by crowdworkers and contains 660 stories, each accompanied by four questions and four possible answers was created in 2013. The complexity of MCTest lies in its reliance on commonsense reasoning and the need to synthesize information across multiple sentences. The SQuAD~\cite{rajpurkar2016squad} is one of the most well-known MRC datasets. It offers a large number of question-answer pairs derived from Wikipedia articles, providing a rich resource for training and evaluating MRC systems. The CoQA dataset~\cite{reddy2019coqa} contains 127k questions with answers, obtained from 8k conversations about text passages from seven diverse domains. Each dialogue in CoQA contains multiple questions and answers, providing a rich resource for training and evaluating QA systems. The SciMRC dataset~\cite{zhang2023scimrc} includes perspectives designed for beginners, students, and experts. Having been constructed from 741 scientific papers it contains 6,057 question-answer pairs where the different perspectives (questions for beginners, students, and experts) contain 3,306, 1,800, and 951 QA pairs, respectively. 

In the field of Arabic Machine Reading Comprehension (MRC), various approaches have been used to tackle different types of questions. For instance,~\citet{azmi2016answering} worked on answering “why” questions using classic Information Retrieval (IR) methods and rhetorical structure theory. Their work was evaluated on a set of 100 questions, demonstrating the application of traditional methods in MRC. The DefArabicQA~\cite{trigui2010defarabicqa} focuses on definition questions and employs an answer ranking module based on word frequency. QArabPro~\cite{akour2011qarabpro} represents a significant step in Arabic MRC, using a rule-based system and achieving high accuracy on a set of questions from Wikipedia. The SemEval tasks of 2015, 2016, and 2017 introduced community QA in Arabic, involving tasks like ranking paragraphs, questions, and multiple answers in order of relevance. Arabic SQuAD~\cite{mozannar2019neural}, a machine-translated version of the English SQuAD dataset, has become a key resource in Arabic MRC, despite the challenges posed by the translation. With its large number of questions and wide coverage of topics, it represents a significant step in developing neural machine reading comprehension benchmarks for Arabic.

In this context, the objective of our paper is to develop ArabicaQA, a comprehensive, large-scale, crowd-sourced dataset for Arabic QA. This dataset, derived from a diverse collection of Standard Arabic documents, aims to foster the advancement of MRC and Open-Domain QA systems for Arabic language resources. Furthermore, ArabicaQA is designed to use with the advancements in LLMs, offering a robust platform for benchmarking and enhancing these models specifically for the Arabic language. In addition to ArabicaQA, we introduce AraDPR, a novel dense passage retrieval model tailored for Arabic text. AraDPR addresses specific challenges inherent to Arabic NLP, marking a significant stride in Arabic text retrieval. This model, when used in conjunction with ArabicaQA, provides a complete solution for Arabic question answering and information retrieval, pushing the boundaries of current capabilities. In Table \ref{tab:dataset}, we highlight the differences between ArabicaQA and other related datasets.
\footnotetext[1]{\textsuperscript{†}Human annotated}
\footnotetext[2]{\textsuperscript{‡}Converted from MRC to Open-domain using an automatic script: \url{https://github.com/deepset-ai/haystack/blob/v1.24.x/haystack/utils/squad_to_dpr.py} It leverages the Haystack library to facilitate the conversion, utilizing Elasticsearch or FAISS for document storage and BM25 or Dense Passage Retriever for retrieving contexts. }
\begin{figure*}[t!]
\centering
\begin{subfigure}[b]{\linewidth}
    \centering
    \includegraphics[width=\linewidth]{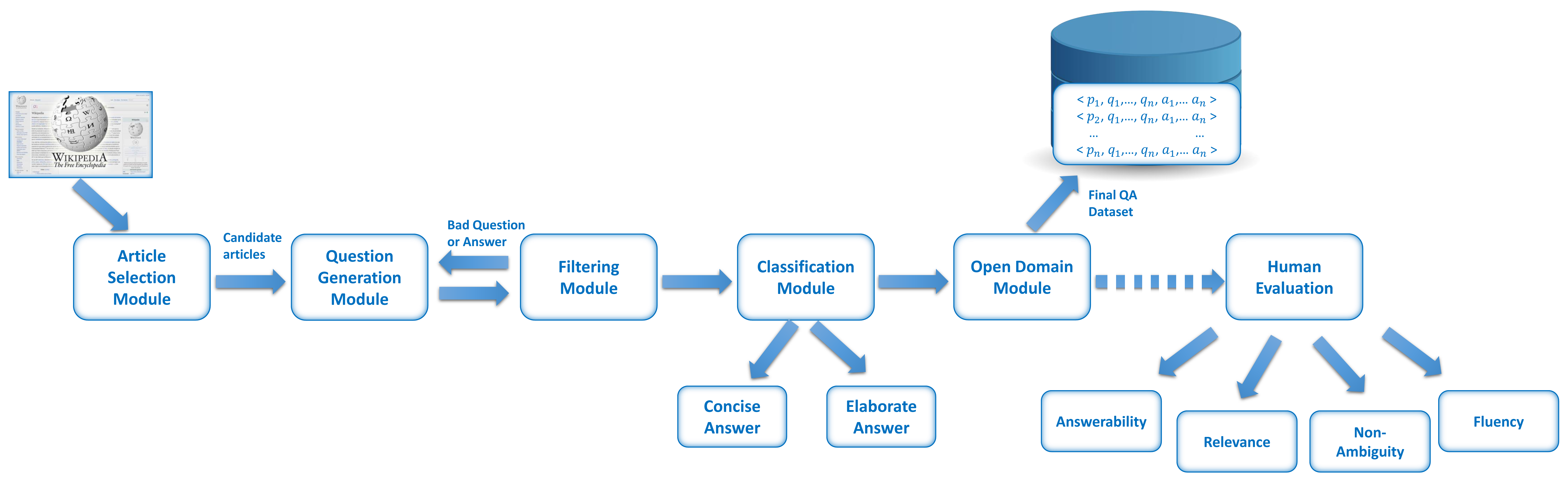}
\end{subfigure}
\caption{The workflow of the QA Dataset Generation Framework integrates human expertise across its stages. Starting with the Article Selection Module, where experts select relevant articles, the process moves to Question Generation, with human-crafted QA pairs. The Filtering Module then involves expert review to discard low-quality pairs, followed by the Classification Module where pairs are categorized into 'Concise Answer' and 'Elaborate Answer' types based on human assessment. In the Open Domain Module, these curated pairs are compiled into the final QA dataset, which later undergoes a Human Evaluation phase for assessing answerability, relevance, clarity, and fluency.}
\label{fig:workflowqa}
\end{figure*}

\section{Methodology}
In this section, we detail the methodology used to create ArabicaQA.
The development process is divided into five primary modules: Article Selection, Question Generation, Filtering/Transforming, Human Answer Categorization, and ODQA Annotation Module. These modules, illustrated in Figure \ref{fig:workflowqa}, were meticulously designed to generate high-quality, relevant question-answer pairs in Modern Standard Arabic that mirror real-world scenarios and information-seeking behaviors.

\subsection{Article Selection Module} The creation of our dataset, ArabicaQA, starts with the selection of source articles. We chose articles from Arabic Wikipedia as our primary source.
The articles were chosen randomly to prevent bias towards certain topics. Each article was examined carefully to ensure it adhered to our length criteria, with a minimum of 20 words and a maximum of 500 words per paragraph. This word limit was set to provide enough context for generating questions while preserving the focus and coherence of the content. We employed 4 proficient Arabic-speaking crowdworkers for this task. Each crowdworker, who is proficient in Arabic, was allocated a subset of these articles. The crowdworkers' responsibility was to generate question-answer pairs based on the paragraphs, ensuring that every question was promptly grounded in the provided content.

\subsection{Question Generation Module}

Question generation is the base of our dataset creation process. To maintain high standards, crowdworkers were provided with a detailed structure and guidelines for creating questions. We employed 10 Arabic-speaking crowdworkers for this task. These guidelines emphasized the importance of generating fact-seeking, clear, and unambiguous questions that could be answered with entities (people, organizations, locations, etc.) or explanatory information directly from the text.

Key aspects of the guidelines included:
\begin{enumerate}
    \item \textit{Clarity and Precision:} Questions should be understandable without additional interpretation and should avoid any assumptions.
    \item \textit{Reformulation:} Avoid copying text verbatim from the passage. Questions should use synonyms and varied sentence structures to ensure they are distinct from the answer text.
    \item \textit{Natural Language Use:} Questions should reflect natural language inquiry as if asking another person for information.
    \item \textit{Answer Formatting:} Concise answers are meant to be short and to the point, while Elaborate answers could include whole sentences or text passages, but not excessively long (no more than 8-10 sentences).
    \item \textit{Inclusion of Imperfections:} Grammatical errors or misspellings, if found, were decided to be included, reflecting the natural variance in user queries.
\end{enumerate}

\subsection{Filtering Module}

The final step in our dataset creation process was the filtering and transforming module, overseen by experts in the Arabic language. This phase involved a thorough review of the generated question-answer pairs. Any questions or answers that were inappropriate, ambiguous, or did not meet the set guidelines were sent back to the annotators for revision.

This iterative process of generation, review, and refinement was crucial in ensuring the quality and reliability of our dataset. It allowed us to remove any inconsistencies, inaccuracies, or biases in questions and answers. 
\begin{figure}[ht]
\fontfamily{cmss}\selectfont
\footnotesize
\centering

\textcolor{cyan}{\rule{\linewidth}{1pt}}
\textbf{Elaborate Answer}
\begin{RLtext}
 - مع أي أندية لعب فرناندو بيليغرينو خلال مسيرته الاحترافية؟ \\
 لعب مع أتلتيكو أتلانتا وأرسنال ساراندي وانيستتوتو وبانفيلد وجيمناسيا لابلاتا وديفنسا خوستيكا وريفر بليت وفيرو كاريل أويستي ونادي ساليرنيتانا \\
\end{RLtext}
\textbf{English Translation:}
\begin{quote}
Q: With which clubs did Fernando Pellegrino play during his professional career? \\
A: He played with Atlético Atlanta, Arsenal de Sarandí, Instituto, Banfield, Gimnasia La Plata, Defensa y Justicia, River Plate, Ferro Carril Oeste, and Salernitana
\end{quote}
\textcolor{cyan}{\rule{\linewidth}{1pt}}
\textbf{Concise Answer}
\begin{RLtext}
في أي مدينة وُلد ساي مين تون؟ \\
-  تاونغيي \\
\end{RLtext}
\textbf{English Translation:}
\begin{quote}
Q: In which city was Sai Min Tun born? \\
A: Taunggyi
\end{quote}
\textcolor{cyan}{\rule{\linewidth}{1pt}}
\caption{Elaborate/Concise answer examples}
\label{fig:examplelongshort}
\end{figure}
\subsection{ Answer Categorization}

An important component of our dataset construction process is the categorization of answers based on an evaluation by language experts. This step ensures that each question is paired with an appropriately categorized, elaborate, or concise answer based on its content and complexity. This classification is important for creating a dataset that can effectively train and evaluate QA systems, particularly those designed for Arabic language processing, in recognizing and generating responses of varying lengths and detailedness levels. Fig. \ref{fig:examplelongshort} shows samples from the elaborate and concise answer questions. 
\paragraph{Elaborate Answer Criteria} After a thorough examination by our Arabic linguistics experts, the dataset includes 39,937 questions categorized under the Elaborate Answer Criteria. These questions are identified for their depth and comprehensiveness, extending beyond mere length to provide a full understanding of the query. This category is representative of questions necessitating detailed explanations or contextual information for satisfactory answers. Our experts' analysis ensures each answer aligns with the question, offering a comprehensive perspective where necessary, thus maintaining the relevance and depth of the responses.
\paragraph{Concise Answer Criteria} Conversely, the dataset features 52,658 questions classified as having concise answers. This categorization is tailored to responses that are succinct and to the point, typically involving entities, dates, names, or other specific details that directly address the query. 
In instances where a concise answer is unfeasible or the question doesn't lend itself to a concise response, annotators are instructed to note the absence of a possible short answer, emphasizing our focus on accurately capturing the essence of the query-response dynamic in Arabic language processing.

\subsection{ODQA Annotation Module}

The Open Domain QA Annotation Module, the final part of the ArabicaQA dataset creation, focuses on annotating broad questions applicable to be asked against a document collection. This process begins with annotators identifying potential open domain questions that span various subjects, followed by conducting internet searches to ensure these questions can be realistically answered with available online information. Each question's answer is verified for accuracy and relevance, undergoing quality assessment for clarity and coherence. Issues identified at any stage require a revision, to maintain the dataset's high standards.

\section{Dataset Analysis}
\subsection{Dataset Statistics}
ArabicaQA is divided into three parts: the training set (70\% of the dataset, with 62,186 answerable and 2,596 unanswerable questions), the development set (15\% of the dataset, with 13,483 answerable and 561 unanswerable questions), and the test set (also 15\% of the dataset, with 13,426 answerable and 544 unanswerable questions).

\begin{figure}[h]
\centering

\begin{subfigure}[b]{\linewidth}
    \centering
    \includegraphics[width=\linewidth]{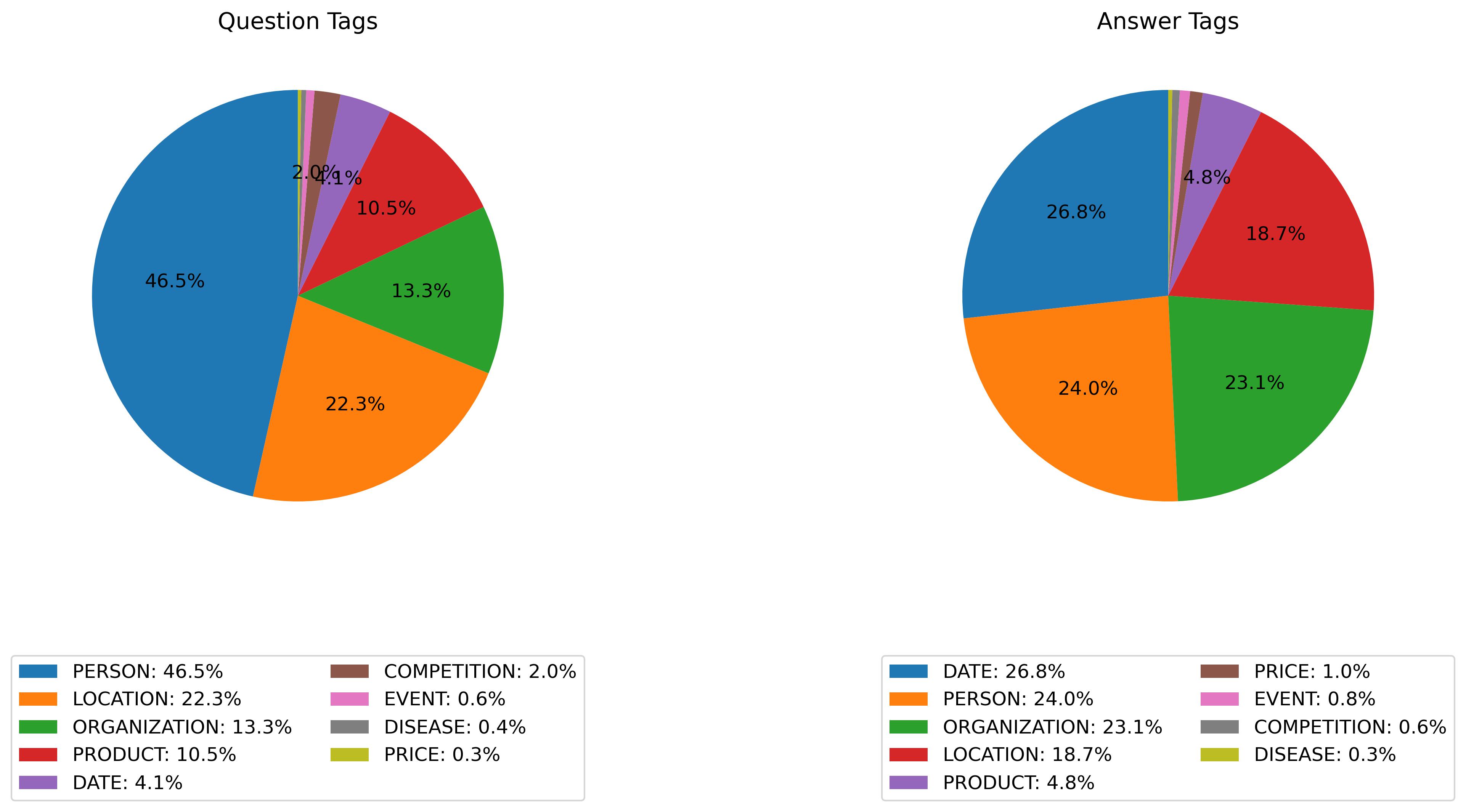}
    \caption{ Questions and Answers NER }
\end{subfigure}%

\caption{Distribution of entity types in the dataset}
\label{fig:entity-distributions}
\end{figure}

\begin{figure}[h]
\centering
    \includegraphics[width=0.7\linewidth]{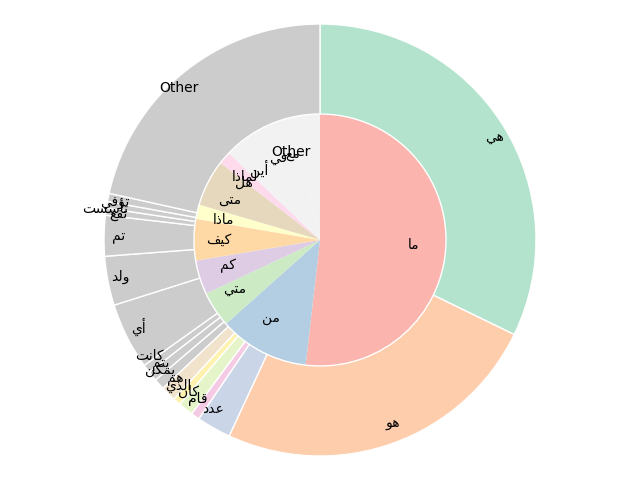}
\caption{Bi-gram Frequencies from ArabicaQA Questions}
\label{fig:gram-distributions}

\end{figure}

For the open-domain tasks, the distribution is slightly different, with 58,676 for training, 12,715 for development, and 12,592 for testing. More detailed dataset statistics are presented in Table \ref{datatset:distribution}.  Also, we conducted named entities analysis for both the questions and answers of our dataset. Named entities refer to specific, identifiable items like dates, names, or locations. Identifying these entities helps in understanding the types of questions our model is capable of answering. As shown in Fig. \ref{fig:entity-distributions}, we observed various entities such as dates, people, organizations, locations, products, competitions, events, diseases, and prices. For the \textbf{answers}, the most common entities are dates, people, organizations, and locations. The most frequently mentioned entities for the \textbf{questions} are people, locations, organizations, and products.

This entity recognition is based on Arabic BERT~\cite{antoun2020arabert}. The NER Arabic BERT was trained on a corpus of 378,000 tokens (14,000 sentences) collected from the Web and manually annotated\footnote{\url{https://huggingface.co/hatmimoha/arabic-ner}}. The results on a validation corpus of 30,000 tokens point to F-1 of~$87$\%.  

Fig. \ref{fig:gram-distributions} also presents the distribution of frequent trigram prefixes. An interesting aspect of our trigram analysis is the prominence of \RL{ما}, which accounts for over $50$\% of the trigram prefixes. In Arabic, \RL{ما} is a multifaceted term that can take on multiple meanings depending on the context and the words that follow it. While \RL{ما} translates to 'what' in English, its usage in Arabic is more complex and varied. 
For instance, in the phrase \RL{ما هو عدد}, it means \textit{how many}, indicating a question about quantity. In another context,  it can be used in a statement such as \RL{ما هي اللغة القديمه للمصريين ؟}, which means \textit{What is the ancient language of the Egyptians?}, It is crucial to note that \RL{ما} in Arabic is context-dependent and does not always directly translate to \textit{what} in English. Its frequent occurrence in our dataset is not just a reflection of its common usage but also highlights the complexity and context-dependent nature of Arabic syntax and semantics.

\subsection{Human Evaluation}
In order to validate the quality of the question-answer pairs generated for our dataset, we followed the human evaluation strategy similar to~\cite{wang2022archivalqa}. We randomly selected 1,000 pairs from our dataset, each accompanied by their original context. Four experts in the Arabic language were then recruited to assess each question-answer pair. The experts, all with proficiency in Arabic linguistics, were instructed to assign ratings on a scale from 1 (very bad) to 5 (very good). The criteria for evaluation were as follows:

\begin{itemize}
    \item \textbf{Fluency:} Assesses the grammatical correctness and readability of a question.
    \item \textbf{Answerability:} Determines if the provided answer correctly answers the question.
    \item \textbf{Relevance:} Evaluates whether the question relates to the content of the given passage.
    \item \textbf{Non-ambiguity:} Judges if the question is formulated unambiguously.
\end{itemize}

Each expert's scores were collected, and an average score for each criterion was computed to reflect the overall evaluation. The results are presented in Table \ref{tab:human_evaluation}.

\begin{table}[h]
    \centering
    \caption{Human Evaluation Scores of the Question-Answer Pairs}
    \label{tab:human_evaluation}
    \begin{tabular}{lcccc}
        \toprule
        Criteria & Fluency & Answerability & Relevance & Non-ambiguity \\
        \midrule
        Expert 1 & 4.908 & 4.735 & 4.730 & 4.754 \\
        Expert 2 & 4.886 & 4.697 & 4.768 & 4.650 \\
        Expert 3 & 4.638 & 4.636 & 4.636 & 4.586 \\
        Expert 4 & 4.699 & 4.150 & 4.638 & 4.157 \\
        \midrule
        Average & 4.782 & 4.554 & 4.693 & 4.536 \\
        \bottomrule
    \end{tabular}
\end{table}

As indicated in Table \ref{tab:human_evaluation} the average scores were commendable across all metrics. The high scores, particularly in fluency and relevance, underline 
high-quality content in our dataset for the Arabic language. Notably, the non-ambiguity scores are also relatively high, suggesting that the majority of the questions were clear and direct.

\subsection{Difficult/Easy Questions}
\begin{figure}[h]
\centering
    \centering
    \includegraphics[width=0.9\linewidth]{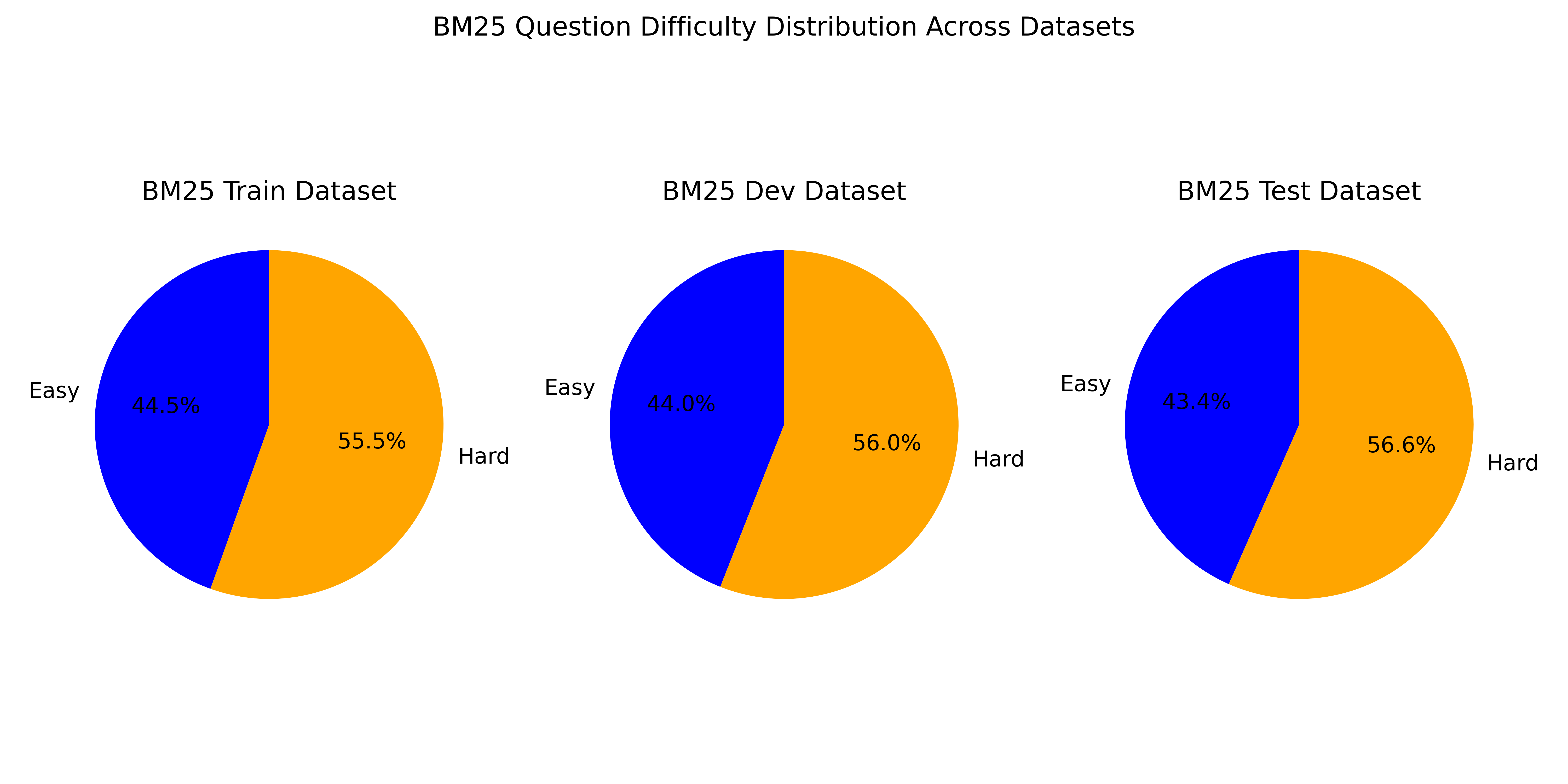}
\caption{Question Difficulty Distribution}
\label{fig:Difficulty}
\end{figure}

We describe how the categorization of questions based on their difficulty levels. The Anserini IR toolkit, an open-source platform, was employed in conjunction with the BM25 ranking function\footnote{\url{https://github.com/castorini/anserini}} from Arabic Wikipedia to facilitate this process. The dataset was divided into two subsets: easy and difficult. Questions are classified as ‘easy’ if the paragraphs used to formulate the questions are found within the top 10 documents retrieved. Conversely, if the paragraphs do not appear within these top 10 documents, the questions are deemed ‘difficult’. Fig. \ref{fig:Difficulty} illustrates the distribution of the dataset, which is ranked using the BM25 method. The breakdown of the dataset is as follows: \textbf{Training set}: 27,640 easy and 34,417 difficult questions.\textbf{Development set}: 5,933 easy and 7,542 difficult questions \textbf{Test set}: 5,820 easy and 7,594 difficult questions.

\section{Models} In this section, we describe the methodologies and models employed in our research. We will discuss baseline models used for MRC, the AraDPR model, and the LLMs. 
\subsection{MRC}
The Term Frequency-Inverse Document Frequency (\texttt{TF-IDF} )~\cite{mozannar2019neural}  Reader is one of our baseline models. It is a simple yet still relatively effective non-learning algorithm that uses 4-gram features to find relevant documents. 
The \texttt{Embedding Reader}~\cite{joulin2016fasttext} is another baseline model we use. Unlike the \texttt{TF-IDF Reader}, \texttt{Embedding Reader} uses the fast-text embeddings\footnote{\url{https://fasttext.cc/docs/en/crawl-vectors.html}}.
It calculates the cosine similarity between the embeddings of documents and queries to find the most relevant documents. 
\texttt{AraBERT}~\cite{antoun2020arabert} is an advanced model we implemented for our MRC approach. AraBERT is a variant of BERT, specifically pre-trained on Arabic text. It leverages the power of transformer-based architectures to understand the context of words in a sentence, making it highly effective for tasks like MRC. 
Multilingual \texttt{RoBERTa}~\cite{liu2019roberta} is another advanced model we use. Like AraBERT, RoBERTa is a variant of BERT, but it is known for its high effectiveness. RoBERTa was trained on more data and for more iterations than BERT, leading to improved performance. It is particularly effective for tasks that require understanding the context of words in a sentence, such as MRC.

\subsection{AraDPR: Dense Retrieval for Arabic}
We focus now on Arabic language retriever for open-domain QA. The model that we built, \texttt{AraDPR}, is based on Bi-encoder framework within a monolingual setting. The efficacy of retrieval models using multilingual transformers has been well recognized, especially considering their high generalization across languages. This capability is critical, as it suggests the probability of leveraging cross-lingual knowledge to enhance retrieval systems in languages such as Arabic, which presents its unique set of linguistic challenges. In the lineage of pre-trained transformer models, \texttt{mBERT} and \texttt{AraBERT} emerged as influential models, benefiting from the concept of multilingual embeddings by being trained on a diverse set of languages. In our methodology, we adapt the architecture of DPR (Dense Passage Retrieval), a seminal model in dense retrieval, same as Karpukhin et al.~\cite{karpukhin2020dense}, to suit the Arabic language. DPR computes embeddings for a query $q$ and a passage $p$ through separate encoders, $E_Q$ and $E_P$, which are typically initialized with English BERT's embeddings. The core of the retrieval mechanism is the computation of a similarity score, $s_{q, p}$, as the dot product of the embeddings:
\begin{align*}
    s_{q, p} \delequal \text{sim}(q, p) = E_Q(q)^T E_P(p).
\end{align*}
DPR is trained with a contrastive loss function, which aims to distinguish between relevant (positive) and non-relevant (negative) passages for a given query:
\begin{align*}
    L(q, p^+, \{p^-\}) = 
    - \log \left( \frac
    { \exp(s_{q, p^+}) }
    { \exp(s_{q, p^+}) + \sum_{p^-} \exp(s_{q, p^-}) }
    \right),
\end{align*}
where $p^+$ denotes the positive passage associated with the query $q$, and the set $\{p^-\}$ comprises multiple negative passages. In retrieval tasks, it is common to have explicit positive examples, while negative examples need to be chosen from a large pool. For instance, in our ArabicaQA dataset, passages relevant to a question might be provided, or they can be identified using the answer. All other passages in the collection, although not explicitly specified, are deemed irrelevant by default. The selection of negative examples, often an overlooked aspect, can be crucial for training a high-quality encoder. We use the top passages returned by BM25\footnote{\url{https://github.com/deepset-ai/haystack}} that do not contain the answer but match most question tokens. This approach ensures a balanced and effective training process for our models.

Given a collection of $M$ text passages, AraDPR aims to index these passages in a low-dimensional, continuous vector space. This indexing facilitates the efficient retrieval of the top $k$ passages most relevant to a given input question, aiding the reader module in real-time comprehension tasks. In our experiments, $M$ represents the entire Arabic Wikipedia, which amounts to approximately $4$ million articles. 
\begin{table*}[ht]
  \caption{Performance Comparison on MRC Tasks}
  \label{tab:mrc}
  \begin{tabular}{ccccccccccc}
    \toprule
    \multirow{2}{*}{Method}  & \multicolumn{5}{c}{Dev} & \multicolumn{5}{c}{Test} \\
    \cline{2-11}
           &     EM & Recall & Precision & F1 & Contains & EM & Recall & Precision & F1 & Contains \\
    \midrule
    
    TF-IDF Reader &  0.08 & 5.347 & 16.480 & 7.039 & 4.357 & 0.128 & 5.317 & 16.38 & 7.09 & 4.18 \\
    Embedding Reader & 0.16 & 8.307 & 14.063 &  8.265 & 5.788 & 0.23 & 8.08 & 13.74 & 7.99 & 5.91 \\
    RoBERTa & 0.29 & 21.32 &  5.31 & 7.50 & \textbf{15.43} & 0.37 & 21.93 &5.41  & 7.67 & \textbf{16.17} \\
    \textbf{AraBERT} & \textbf{4.05} & \textbf{34.82} & \textbf{18.44} & \textbf{19.69} & 14.39 & \textbf{4.507} & \textbf{35.74} & \textbf{18.88} & \textbf{20.42} &  14.83\\
   
    \bottomrule
  \end{tabular}
  \label{table:mrc}
\end{table*}
\subsection{LLM Models}
In this section, we describe the details of the Large Language Models (LLMs) utilized in our research. We discuss four distinct models: \texttt{Llama}, \texttt{Mistral}, \texttt{Mixtral}, \texttt{Falcon}, \texttt{Qwen} and \texttt{GPT3.5}. Each of these models brings unique strengths to our research and has been chosen for specific reasons.

\texttt{Llama}\footnote{\url{https://huggingface.co/meta-llama/Llama-2-7b-chat-hf}}, part of the Llama 2 family of LLMs developed by Meta~\cite{touvron2023llama}, comprises pre-trained and fine-tuned generative text models. These models, specifically optimized for dialogue use cases, have shown superior performance on numerous benchmarks. They employ supervised fine-tuning and reinforcement learning with human feedback~\cite{touvron2023llama} to align with human preferences for helpfulness and safety.  

\texttt{Mistral}\footnote{\url{https://huggingface.co/mistralai/Mistral-7B-Instruct-v0.2}}, released by Mistral AI, is known for its power and efficiency. It surpasses the Llama 2 13B on all benchmarks. ~\cite{jiang2023mistral}. It leverages instruction fine-tuning, where the prompt should be surrounded by [INST] and [/INST] tokens. 

\texttt{Mixtral}\footnote{\url{https://huggingface.co/mistralai/Mixtral-8x7B-Instruct-v0.1}}, another innovation from Mistral AI, is a trained generative Sparse Mixture of Experts~\cite{jiang2023mistral} that outperforms the Llama 2 70B model on most benchmarks. The model leverages up to 45B parameters but only uses about 12B during inference, leading to better inference throughput at the cost of more vRAM. 

\texttt{GPT3.5}\footnote{\url{https://platform.openai.com/docs/models/gpt-3-5}} is a fine-tuned version of the GPT3 model with 175B parameters~\cite{brown2020language}. GPT3.5 focuses on reducing the generation of toxic outputs. It uses 12 stacks of decoder blocks with multi-head attention blocks. The training data for GPT3.5 is a diverse mix of texts from the internet, including books, articles, and websites. The model was trained on data up published before December 2023\footnote{\url{https://help.openai.com/en/articles/8555514-gpt-3-5-turbo-updates}}. 

\texttt{Falcon}\footnote{\url{https://huggingface.co/docs/transformers/main/model_doc/falcon}}  is a class of causal decoder-only models. The largest Falcon checkpoints have been trained on >=1T tokens of text, with a particular emphasis on the RefinedWeb corpus~\cite{refinedweb}. Falcon architecture is modern and optimized for inference, with multi-query attention and support for efficient attention variants like FlashAttention~\cite{dao2022flashattention}.  

\texttt{PPLX}\footnote{\url{https://www.searchenginejournal.com/perplexity-introduces-online-llms-with-real-time-information/502523/}}, introduced by Perplexity, is a new online  LLM that leverages the internet for real-time data. The PPLX models, pplx-7b-online and pplx-70b-online, build on top of the open-sourced mistral-7B and llama2-70B models and have been specifically fine-tuned by Perplexity on diverse, high-quality datasets to optimize for helpfulness and factuality. These models can access the latest information from the internet to provide up-to-date responses.

\texttt{Qwen}\footnote{\url{https://github.com/QwenLM/Qwen}}, proposed by Alibaba Cloud, is a comprehensive language model series that encompasses distinct models with varying parameter counts. It includes Qwen~\cite{qwen}, the base pretrained language models, and Qwen-Chat, the chat models fine-tuned with human alignment techniques. The Qwen model series spans from 1.8 billion parameters to 72 billion parameters. 
\subsection{Evaluation Metrics}

\paragraph{Evaluation of MRC and LLMs:} We assess the performance of QA using several metrics, including exact match (EM), precision, token recall, answer string containment, and F1 score. These are standard measures widely used in QA research. However, given that LLMs often generate verbose answers, many standard QA metrics may not be well-suited for evaluating answer quality. For instance, the exact match will almost always be zero due to the presence of non-ground-truth tokens, and the F1 score will be penalized by other potentially useful tokens. To address this, we use a set of model-agnostic metrics, namely token recall and answer string containment \cite{adlakha:2023:arxiv:evalinstructs,liu:2023:arxiv:longcontextshow,mallen:2023:ACL:whennottrustllms}.

\paragraph{Retriever Evaluation} The evaluation of the retriever component in our research employs the top-K retrieval accuracy metric~\cite{sachan2022improving}. This metric calculates the fraction of questions for which at least one passage in the top-K retrieved passages contains a span matching the human-annotated answer. The formula is given by:
\begin{equation}
    \text{Retriever Accuracy@k} = \frac{\sum(\text{any(Correct@k)})}{\text{\# retrieved documents}}
\end{equation}
\section{Experimental Results}
The results of our experiments on the ArabicaQA dataset provide insights into the performance of various models in the domain of Machine Reading Comprehension (MRC), Language Model in-context learning, and information retrieval.
\subsection{MRC Result} Table \ref{table:mrc} presents a comparison of the performance of different methods on MRC tasks, evaluated on both the development (Dev) and test sets. The methods include \texttt{TF-IDF Reader}, \texttt{Embedding Reader}, \texttt{Roberta}, and \texttt{ARaBERT}. 

\texttt{TF-IDF} Reader and Embedding Reader show a relatively low performance across all metrics. This is expected as these models rely on simpler algorithms compared to deep learning-based models. \texttt{RoBERTa} shows a significant improvement in Exact Match (EM) and F1 scores, indicating its better understanding of context compared to the more basic methods. \texttt{ARaBERT}, specifically tailored for the Arabic language, demonstrates the highest performance across almost all metrics. Its superior understanding of Arabic nuances and context is evident in the results.

\begin{figure}[h]
\centering
    \centering
    \includegraphics[width=0.8\linewidth]{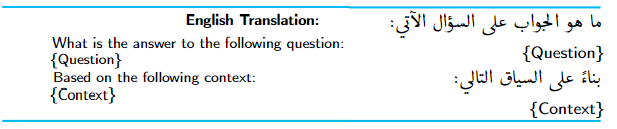}
\caption{Prompt Structure in Arabic and  English Translation}
\label{fig:prompt}
\end{figure}

\label{section:incontext}
In Table \ref{tab:mrcllm}, the performance of the LLMs 
is evaluated in terms of in-context learning for MRC tasks. Falcon shows remarkable performance in terms of EM and Contains scores. Its high capacity (180B parameters) plays a significant role in achieving this result. \texttt{GPT3.5} and \texttt{PPLX} display moderate results, with \texttt{PPLX} showing a notably high Recall but lower EM and Precision. \texttt{Qwen}, \texttt{LLama2}, \texttt{Mistral}, and \texttt{Mixtral} demonstrate varying degrees of performance, with none dominating across all metrics. This indicates that the model's size (in terms of parameters) matters. In Fig. \ref{fig:prompt}, the structure of the prompt is presented in both Arabic and English.  In the prompt, '{Context}' is replaced with a passage of text, and '{Question}' is a query related to that passage.

\begin{table}[h]
  \caption{In-Context Learning with LLMs for MRC}
  \label{tab:mrcllm}
  \begin{adjustbox}{width=0.5\textwidth,center}
  \begin{tabular}{ccccccc}
    \toprule
    \multirow{2}{*}{Method}  & \multirow{2}{*}{Parameter}&\multicolumn{5}{c}{Test} \\
    \cline{3-7}
    &   &  EM & Recall & Precision & F1 & Contains  \\
    \midrule
    Mistral & 7B & 0.067 & 34.80 & 6.110 & 9.311 & 22.28   \\
    Qwen & 14B & 0.185 & 15.81  &6.86  & 8.533  &  7.862  \\
    Mixtral & 8x7B  &  0.078  & 33.23 &  7.228 & 10.39  &   14.00       \\
    LlaMA2 & 70B & 0.0 & 16.80 & 2.233 & 16.80 & 7.11   \\
     PPLX & 70B &  0.014 & 69.52 & 13.06 & 20.17 &  \textbf{40.23}  \\
     GPT3.5 & 175B & 0.164 &  32.16 &  \textbf{24.30} &  \textbf{32.16} &   32.161\\
    Falcon  & 180B &\textbf{0.613} & \textbf{46.94} & 15.57 & 20.93 & 24.79   \\
    
    \bottomrule
  \end{tabular}
    \end{adjustbox}
\end{table}

\begin{table*}[t]
  \caption{Retriever Module Performance Comparison}
  \label{tab:Retrieveracc}
  \begin{tabular}{cccccccccccc}
    \toprule
    \multirow{2}{*}{Method} & \multirow{2}{*}{Type} & \multicolumn{5}{c}{Dev} & \multicolumn{5}{c}{Test} \\
    \cline{3-12}
           &      & 1 & 10 & 20 & 50 & 100 & 1 & 10 & 20 & 50 & 100 \\
    \midrule
    TF-IDF & Unigram &  10.96 &  30.25 & 37.68 &  47.87 & 55.78 &10.33  & 29.11 & 36.09 & 46.42 & 54.68 \\
     TF-IDF & Hierarchical & 10.97  & 30.44 & 37.27 &  47.91 & 56.48 &10.43  & 29.73 & 37.00 & 47.08 &55.77  \\
     TF-IDF & Bigram & 13.80 & 39.73 & 48.96 & 51.34 & 55.87 & 14.35 &   40.86 &46.87&51.71& 55.36\\
     BM25& -& 27.60 & 44.10 & 49.20 & 55.70 & 60.60 & 28.70 & 43.40 & 48.20 &54.60  & 59.30   \\
     DPR & multi-language BERT & \textbf{36.50} & 58.60 & 62.70 & 67.20 & 70.20 &  \textbf{36.40} & 57.80 & 62.10 & 66.60 & 69.50  \\
      AraDPR & AraBERT& 36.30 & \textbf{59.60} & \textbf{64.10} & \textbf{69.30} & \textbf{72.50} &  36.10 & \textbf{58.40} & \textbf{63.40} &  \textbf{68.60} & \textbf{71.90} \\

    \bottomrule
  \end{tabular}
\end{table*}
\subsection{Retriever Results}

The performance of different retrieval methods is presented in Table \ref{tab:Retrieveracc}, with methods including various \texttt{TF-IDF } configurations,  \texttt{BM25}, and two versions of  \texttt{DPR} (multi-language  \texttt{BERT} and  \texttt{AraBERT}). \texttt{TF-IDF} in its various forms (Unigram, Hierarchical, Bigram) shows the worst performance. 
The Bigram model generally outperforms the Unigram and Hierarchical models. \texttt{BM25} outperforms \texttt{TF-IDF} methods, especially at lower values of K, which suggests its effectiveness in quickly identifying relevant documents. \texttt{DPR} with multi-language \texttt{BERT} and  \texttt{AraBERT} both show superior performance over the other methods, with  \texttt{AraBERT} slightly leading. This indicates the effectiveness of dense retrieval methods and the advantage of language-specific tuning (AraBERT).

\subsection{Retrieval-augmented Generation}
Retrieval-Augmented Generation (RAG)~\cite{lewis2020retrieval,abdallah2023generator}, merges the capabilities of advanced language models and information retrieval systems for response generation in question-answering scenarios. In this approach, upon receiving a question, pertinent documents are initially sourced from an extensive corpus. Subsequently, these documents serve as a supplementary context for a language model. This model then crafts an answer, taking into account both the initial query and the information from the retrieved documents. In this experiment, we aim to evaluate the efficiency of appending the retrieved context to LLMs for question answering. Table \ref{tab:freq} evaluates the performance of LLMs (\texttt{GPT3.5}, \texttt{Mixtral}, \texttt{LLama}, \texttt{Mistral}, \texttt{Qwen}) when provided with the first retrieved document for Open-domain QA tasks. \texttt{GPT3.5} shows moderate performance, with a balance across all metrics. \texttt{Mixtral} and \texttt{LLama} have lower scores, indicating a potential limitation in utilizing the first retrieved document for generating accurate responses. \texttt{Mistral} and \texttt{Qwen} show somewhat better performance, particularly in Recall and Contains scores, suggesting their capability to use the retrieved documents effectively. In this experiment, we used the same prompt as in Section \ref{section:incontext}

\begin{table}[h]
  \caption{Retrieval-augmented Generation with first retrieved document Result}
  \label{tab:freq}
  \begin{adjustbox}{width=0.5\textwidth,center}
  \begin{tabular}{ccccccc}
    \toprule
    \multirow{2}{*}{Method}  & \multirow{2}{*}{Parameter}&\multicolumn{5}{c}{Test} \\
    \cline{3-7}
           &   &  EM & Recall & Precision & F1 & Contains  \\
    \midrule
    Mistral & 7B & 0.0 &24.46 & 5.35& 8.01 &8.47  \\
    Mixtral & 8x7B  &0.03& 8.33 &2.03 & 2.94 & 2.46 \\  
    LLama & 70B  & 0.0 & 9.63 & 1.45 & 2.35 &  4.10 \\
    Qwen &14B & 0.25 & 29.94 & 12.72 & 16.08 & 15.17 \\
    \textbf{GPT3.5}  & \textbf{175B}  & \textbf{0.037} & \textbf{36.80} & \textbf{13.77} & \textbf{18.32} & \textbf{17.53} \\

    \bottomrule
  
  \end{tabular}
    \end{adjustbox}
\end{table}
\section{Dataset Use}
ArabicaQA is a versatile resource for Natural Language Processing, especially for tackling Arabic's complex linguistic features in Machine Reading Comprehension and Open-Domain Question Answering. Its diverse and extensive content aids in developing models that grasp Arabic's rich morphology and syntax. ArabicaQA is instrumental for improving Information Retrieval systems, and enhancing document relevance through a deep understanding of context, making it a critical tool for advancing Arabic NLP technologies. ArabicaQA could enhance passage ranking algorithms. This is particularly crucial for Arabic, where the relevance of retrieved documents is heavily influenced by contextual nuances.


Moreover, ArabicaQA is an ideal resource for training models in question generation tasks. This field has seen limited progress in Arabic due to the lack of specialized datasets~\cite{al2023challenges,alsaleh2021snad,boujou2021open}. By providing a substantial volume of question-answer pairs, our dataset can contribute to advancing research in this area as well. The development of NLP models for Arabic is fraught with challenges. Arabic’s rich morphology results in a high degree of word form variability. This, coupled with the language’s reliance on context for meaning, makes accurate comprehension a formidable task for current models. Our dataset, with its diverse range of texts and contexts, is specifically designed to address these challenges. It provides a varied and rich corpus for training models that can navigate the complexities of Arabic syntax and semantics. Given the detailed and specific nature of our dataset, it also holds potential for educational applications, particularly in teaching and evaluating knowledge in Arabic language courses. The diverse content of the dataset can be used to stimulate self-learning of students, especially in understanding the nuances and complexities of Arabic. 

To sum up, ArabicaQA stands as a scalable and substantial resource for researchers aiming to contribute to Arabic language processing. The lack of comprehensive and challenging datasets has been a major bottleneck in Arabic NLP research. By introducing a dataset that is both extensive and tailored to the unique aspects of Arabic. This is especially crucial for developing models that can handle the complexities of Arabic, thereby contributing to the broader goal of creating truly global and inclusive NLP technologies.

\section{Conclusion}
This paper introduces ArabicaQA, a large dataset designed to boost the development of Arabic NLP, specifically MRC and open-domain QA. The dataset is unique, containing both answered and unanswered questions, as well as answers that are concise or elaborate, making it a challenging and enriching resource for testing and improving NLP models. ArabicaQA has been manually created and is currently the largest dataset for Arabic QA. Our study evaluated various models on the ArabicaQA and found that advanced models like AraBert and Roberta, which are fine-tuned for Arabic, outperform traditional models. Large Language Models (LLMs) like Falcon, GPT3.5, and PPLX showed varying success levels in in-context learning for MRC tasks, highlighting the complexity of language processing and the need for model-specific tuning. 

Our research also confirmed that dense retrieval methods, such as DPR models, are more effective than traditional retrieval methods. However, LLMs face challenges in using retrieved documents to improve MRC accuracy, suggesting a need for further research in this area. In conclusion, ArabicaQA is a significant contribution to Arabic NLP research, providing valuable resources and insights. Our hope is that this work will inspire further research and development in Arabic NLP leading to more advanced and efficient language processing systems.

\textbf{Limitations} We acknowledge the following limitations of our work. ArabicaQA is focused on Modern Standard Arabic and is sourced from Wikipedia, which limits its linguistic diversity and domain coverage. The dataset might also benefit from incorporating a broader range of question complexities and addressing potential annotation biases.
Annotation biases introduced through crowd-sourcing could also impact the dataset's neutrality and quality, despite rigorous review processes. 
Addressing these limitations forms a part of our future work.
\newpage

\bibliographystyle{ACM-Reference-Format}
\bibliography{sample-base}

\end{document}